%% file: main.tex
\documentclass[sigconf]{acmart}
\AtBeginDocument{%
  }

\setcopyright{acmlicensed}
\copyrightyear{2018}
\acmYear{2018}
\acmDOI{XXXXXXX.XXXXXXX}

\acmConference[Conference acronym 'XX]{Make sure to enter the correct
  conference title from your rights confirmation emai}{June 03--05,
  2018}{Woodstock, NY}
\acmISBN{978-1-4503-XXXX-X/18/06}

\usepackage{booktabs}
\usepackage{multirow}
\usepackage{makecell}
\usepackage{subfigure}
\usepackage{bm}
\usepackage{amsmath}
\usepackage{xcolor}
\usepackage[normalem]{ulem}
\useunder{\uline}{\ul}{}

\begin{document}


% \title{Disentangled Multi-layer Evolutionary Graph Neural Network for Temporal Link Prediction}
% \title{Disentangled Multi-layer Evolutionary Graph Neural Network against Temporal Link Prediction}
% \title{Disentangled Multi-layer Evolutionary Neural Network against Temporal Knowledge Graph Reasoning}
\title{Disentangled Multi-span Evolutionary Network against Temporal Knowledge Graph Reasoning}
% \title{Disentangled Cross-time Layer Evolutionary Network against Temporal Knowledge Graph Reasoning}
% \title{Multi-layer Evolutionary Graph Neural Network for Temporal Link Prediction}
% \title{Multi-layer Evolutionary GNN for Temporal Link Prediction}

\author{Hao Dong}
\affiliation{%
  \institution{Computer Network Information Center, Chinese Academy of Sciences}
  % \institution{University of Chinese Academy of Sciences}
  \city{Beijing}
  \country{China}
}
\email{donghcn@gmail.com}

\author{Ziyue Qiao}
\affiliation{%
  \institution{School of Computing and Information Technology, Great Bay University}
  \city{Dongguan}
  \country{China}
}
\email{ziyuejoe@gmail.com}

\author{Zhiyuan Ning}
\affiliation{%
  \institution{Computer Network Information Center, Chinese Academy of Sciences}
  % \institution{University of Chinese Academy of Sciences}
  \city{Beijing}
  \country{China}
}
\email{zyningcn@gmail.com}

\author{Qi Hao}
\affiliation{%
  \institution{State Key Laboratory of Internet of Things for Smart City, University of Macau}
  \city{Macau SAR}
  \country{China}
}
\email{yc37440@um.edu.mo}

\author{Yi Du}
\affiliation{%
  \institution{Computer Network Information Center, Chinese Academy of Sciences}
  % \institution{University of Chinese Academy of Sciences}
  \city{Beijing}
  \country{China}
}
\email{duyi@cnic.cn}

\author{Pengyang Wang}
\authornote{Corresponding author.}
\affiliation{%
  \institution{State Key Laboratory of Internet of Things for Smart City, University of Macau}
  \city{Macau SAR}
  \country{China}
}
\email{pywang@um.edu.mo}

\author{Yuanchun Zhou}
\affiliation{%
  \institution{Computer Network Information Center, Chinese Academy of Sciences}
  % \institution{University of Chinese Academy of Sciences}
  \city{Beijing}
  \country{China}
}
\email{zyc@cnic.cn}

\renewcommand{\shortauthors}{Dong et al.}

\input{0.abs}




\received{20 February 2007}
\received[revised]{12 March 2009}
\received[accepted]{5 June 2009}

\maketitle

\input{1.intro}

\input{3.pre}

\input{4.method}
\input{5.exp}

\input{2.rel}
\input{6.con}

\begin{acks}
This research was supported by 
the National Natural Science Foundation of China (No.92470204),
the Science and Technology Development Fund, Macau SAR (file no. 0123/2023/RIA2, 001/2024/SKL),
the National Natural Science Foundation of China (No. 62406056),
and the Youth Innovation Promotion Association CAS.
\end{acks}

\bibliographystyle{ACM-Reference-Format}
\bibliography{ref}

\end{document}

%% file: 0.abs.tex
\begin{abstract}
Temporal Knowledge Graphs (TKGs), as an extension of static Knowledge Graphs (KGs), incorporate the temporal feature to express the transience of knowledge by describing when facts occur.
TKG extrapolation aims to infer possible future facts based on known history, which has garnered significant attention in recent years. 
Some existing methods treat TKG as a sequence of independent subgraphs to model temporal evolution patterns, demonstrating impressive reasoning performance.
However, they still have limitations:
1) In modeling subgraph semantic evolution, they usually neglect the internal structural interactions between subgraphs, which are actually crucial for encoding TKGs.
2) They overlook the potential smooth features that do not lead to semantic changes, which should be distinguished from the semantic evolution process.
Therefore, we propose a novel \textbf{\underline{Di}}sentangled \textbf{\underline{M}}ulti-span Evolutionary \textbf{\underline{Net}}work (\textbf{DiMNet}) for TKG reasoning. 
Specifically, we design a multi-span evolution strategy that captures local neighbor features while perceiving historical neighbor semantic information, thus enabling internal interactions between subgraphs during the evolution process.
To maximize the capture of semantic change patterns, we design a disentangle component that adaptively separates nodes' active and stable features, used to dynamically control the influence of historical semantics on future evolution.
% Additionally, we design an inference scheme specifically for the multi-span evolution encoding to mitigate the challenges posed by the uncertainty of future distribution.
Extensive experiments conducted on four real-world TKG datasets show that DiMNet demonstrates substantial performance in TKG reasoning, and outperforms the state-of-the-art up to 22.7\% in MRR.
\end{abstract}

%% file: 1.intro.tex
\section{Introduction}

Knowledge Graphs (KGs) widely serve to represent structured facts about the real world. They are applied in various intelligent domains, such as information retrieval~\cite{gaur2022iseeq}, knowledge-based question answering~\cite{lan-jiang-2020-query, 10.1016/j.ins.2022.11.042}, recommendation systems~\cite{10.1145/3292500.3330989, 10.1007/s10489-021-02872-8}, traffic prediction~\cite{zhou2024make}, etc. Typically, to represent the topological structure of the static entities in facts, traditional KGs integrate and store static knowledge in the form of triples $(s,r,o)$, which consist of subject, relation, and object.

However, factual knowledge and entity semantics are constantly evolving over time, exhibiting complex temporal characteristics, which inspired recent literature to propose Temporal Knowledge Graph (TKG) to represent facts~\cite{dong2023adaptive, dong2024temporal}. Specifically, it introduces a time feature to record the timestamps or time intervals related to events and takes quadruples to represent a temporal fact, denoted as $(s,r,o,t)$, such as $(Thomas\ Alva\ Edison, BornIn, USA, 02/11/1847)$. TKGs have a broader range of applications, such as crisis warning~\cite{liu2022article} and real-time dialogue~\cite{liu2021lifelong}. 

Similar to static KGs, TKGs also face the challenge of incompleteness. The task of reasoning on TKGs aims to complete missing links based on given known temporal facts. This includes two settings: interpolation and extrapolation~\cite{jin2020Renet}. 
The former is used to predict missing facts within a known historical time range, while the latter predicts future facts based on known historical facts~\cite{li2021temporal}. In this work, we focus on extrapolation reasoning that presents significant challenges and a wide scope for investigation.

% Describe how other literature solve the extrapolation task

To accurately infer future facts, it is crucial to comprehensively understand known facts in history and capture the patterns in which these historical facts occurred. A few early works start from the future query and extract relevant semantic information and occurrence patterns from history. RE-NET~\cite{jin2020Renet} is a representative work that encodes historical facts related to the query by RNN to obtain query-related representations. CyGNet~\cite{zhu2021learning} directly captures repetitive patterns of facts that share the same entities and relations as the query. Although these methods can make the inference distribution at the prediction stage closer to the future, they overlook the structural information and relative temporal characteristics within the historical facts, which are crucial for TKG extrapolation.
% , leading to an over-reliance on repetitive patterns in historical data during TKG extrapolation.

To make more comprehensive use of historical information and deeply capture the developmental patterns of history, several subgraph evolution methods have been proposed, such as RE-GCN~\cite{li2021temporal}, RETIA~\cite{liu2023retia}, DaeMon~\cite{dong2023adaptive}, etc. They view the TKG as a sequence of subgraphs and use GNN-based methods to model local structural dependencies in history, while employing sequence modeling to capture global semantic evolution patterns. However, this approach results in the structural modeling of the constructed subgraphs being independent of each other, hindering the mutual influence of internal structures between subgraphs.
% ---------above intro the weakness of evolution, below intro the weakness of whole graph
TITer~\cite{2021TimeTraveler} connects the sequence of historical subgraphs into a whole by introducing auxiliary edges. TiPNN~\cite{dong2024temporal} and xERTE~\cite{han2020explainable} sample the historical facts of the TKG into a single comprehensive graph, utilizing time encoding to capture the dynamic characteristics of facts. Although these methods break the isolation of subgraph internal structures, they neglect the features of the semantic evolution pattern of the TKG.

However, the \textit{internal structural interaction} among subgraphs and the \textit{subgraphs' semantic evolution} over sequence are both crucial.
Therefore, \textbf{it still remains challenges: }

$(1)$  Existing models only focus on one of them, which limits their performance on TKG reasoning.
% When we divide the TKG into a sequence of subgraphs based on timestamps, each subgraph represents a snapshot. 
From the viewpoint of GNN's message passing~\cite{scarselli2008graph,ning2021lightcake,ning2025deep}, the semantic evolution over time essentially involves the influence of neighbors in different subgraphs on the central node, leading to changes in the central node's semantics.
During the evolutionary process, to enable beneficial interactions among subgraph structures, the most direct method is to allow the central node to perceive its historical neighbors.
% when aggregating current neighbor information.
However, for a central node, it has neighbors at different distances, and the importance of these neighbors to the central node is distance-related. Therefore, when the central node perceives the historical neighbors' features, the distance between the central node and its neighbors in the historical subgraph should also be taken into account.
% Therefore, when the central node aggregates neighbor message of a certain distance, the historical neighbors information it perceives should also be within the same span.
% $(2)$ Semantic evolution is a high-order implicit modeling process. However, 
% $(2)$ Each node has its inherent intrinsic properties. These intrinsic properties remain relatively stable during the process of semantic evolution, which has also been overlooked by existing methods.

$(2)$ When the neighbor changes along the timeline, the semantics of a node evolve. However, the changes in the neighbor may not be entirely thorough, and each central node also has its inherent attributes. 
Subgraph semantic evolution is a higher-order implicit modeling process. During the evolution, there are stable semantics of nodes that change relatively smoothly. Although~\cite{li2021temporal} introduced static entity types in evolutionary modeling, which is not a generalizable method since the duration of stable semantics is often unpredictable. 
For subgraph evolution, features with stable changes should be distinguished so that the model can maximize the capture of changes in node semantic information.

% To address these challenges, we start with a fundamental perspective: "\textit{When GNN Meets Temporal Knowledge Graph}". 

To this end, we propose the \textbf{\underline{Di}}sentangled \textbf{\underline{M}}ulti-span Evolutionary \textbf{\underline{Net}}work (\textbf{DiMNet}) based on an encoder-decoder structure for the TKG reasoning task. 
The encoder models the evolutionary process of the historical subgraph sequence, and the decoder infers future facts from the evolutionary results.
Specifically, in the encoding phase of our model, we propose \textbf{Multi-span Evolution} approach based on graph neural network to enable interaction between the internal structures of subgraphs while preserving the modeling of semantic evolution across historical subgraphs. 
It is designed to perceive the semantic information of historical neighbors with equal spans during the local subgraph structural modeling stage, thus realizing the perception of the distance features of historical neighbors
% and guide the structural modeling process of the current subgraph 
(First challenge).
Additionally, during the subgraph evolution process, we design a \textbf{Disentangle Component} to dynamically and adaptively separate the mutually exclusive active and stable features of nodes as they evolve over time. 
The stable feature is used to guide the node evolution without deviating from the node's steady-state characteristics and intrinsic properties. 
The active feature is to guide the influence of multi-span historical neighbors on the current subgraph, thereby maximizing the modeling of change patterns (Second challenge).

In general, this paper makes the following contributions:
\begin{itemize}
    \item  We propose a novel method DiMNet for TKG extrapolation, which models the evolution of historical subgraph sequences from the perspective of semantic change. Through a multi-span evolution strategy and a disentangle component, DiMNet is able to learn node semantic change patterns in a fine-grained manner.
    \item We design an inference strategy based on sampling virtual subgraphs specifically for the multi-span evolution encoding method to mitigate the issue of future topology uncertainty during the inference phase.
    \item Extensive experiments on four benchmark datasets demonstrate the effectiveness of our proposed method, showing superior performance compared to baselines for the TKG reasoning task and achieving new state-of-the-art results.
\end{itemize}

%% file: 3.pre.tex
\section{Preliminaries}

Here, we introduce the background of TKG and provide a formal statement.
Then, we present a formulation of the reasoning task. 
% The details of essential mathematical symbols and the corresponding descriptions of TiPNN are shown in Table \ref{tab:notation}. 
Note that, except for special clarifications, we use \textbf{bold} items to denote vector representations in the following context.

\subsection{Definition of TKG}

A TKG $\mathcal{G}$ is composed of several quadruples $(s,r,o,t)$. 
From a temporal perspective, a TKG can be formalized as a sequence of static subgraphs arranged in chronological order, i.e., $\mathcal{G}=\{G_1, G_2,..., G_t,...\}$. 
Each $G_t$ in $\mathcal{G}$ is essentially a directed multi-relational graph at timestamp $t$. It can be represented as $G_t=(\mathcal{V}, \mathcal{R}, \mathcal{E}_t)$, where $\mathcal{V}$ is the set of nodes, $\mathcal{R}$ is the set of relation types, and $\mathcal{E}_t$ is the set of edges at timestamp $t$.
Each element in $\mathcal{E}_t$ can be expressed as $(s_t,r_t,o_t)$, describing a relation type $r \in \mathcal{R}$ occurring between the subject node $s \in \mathcal{V}$ and the object node $o \in \mathcal{V}$ at timestamp $t \in \mathcal{T}$, where $\mathcal{T}$ denotes the finite set of timestamps.

\subsection{Formulation of Reasoning Task}

In this work, we concentrate on the extrapolation reasoning of TKG, which involves completing facts that occur in the future. For example, given an object node query $(s,r,?,t_q)$ at a future timestamp $t_q$, we derive the reasoning results by considering all historical known facts $\{(s,r,o,t_i) | t_i < t_q\}$.
Without loss of generality, inverse edges $(o,r^{-1},?,t_q)$ are also added to the TKG. This allows us to transform the query when predicting the subject, i.e., $(?,r,o,t_q)$, into the form of predicting the object $(o,r^{-1},?,t_q)$. Therefore, for convenience, we will introduce reasoning from the perspective of predicting the object in the following context.

It is worth noting that, similar to previous work, we consider the historical subgraphs corresponding to the most recent $m$ timestamps when performing extrapolation reasoning. In other words, when the future query time is $t+1$, we take the historical subgraph sequence $\{G_{t-m+1},..., G_t\}$ as the known historical information. Additionally, the facts at the query timestamp are completely unknown in this task.

%% file: 4.method.tex
\begin{figure}[!t]
    \centering
    \includegraphics[width=0.47\textwidth]{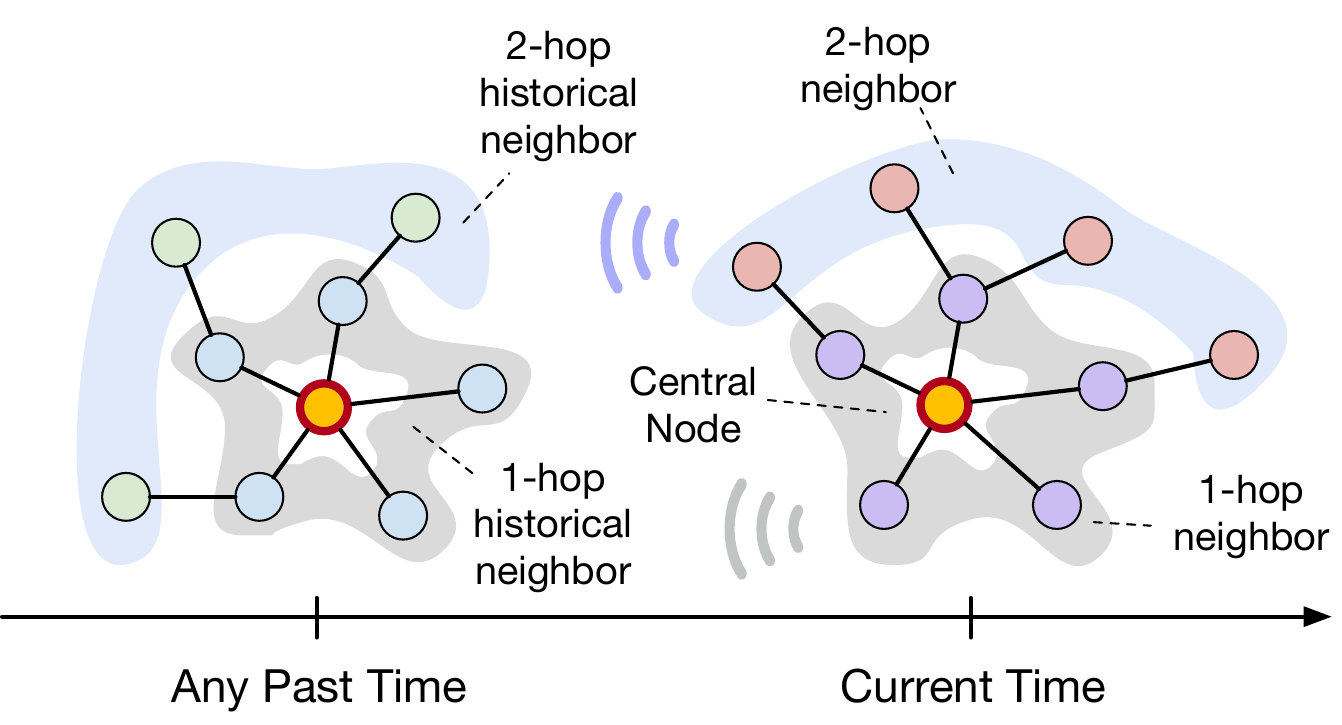}
    \caption{Illustration of the current central node perceiving historical neighbors. When the central node in the current subgraph aggregates neighbor information, historical neighbor information with the same hop is perceived to enhance the semantic aggregation of neighbors.}
    \label{fig:historical-neighbor}
\end{figure}

\section{Methodology}
% In this section, we introduce the proposed model, Disentangled Multi-span Evolutionary Network (DiMNet). We begin with an overview of the model, followed by a discussion of each of its components and the training process.

\subsection{Model Overview}

\begin{figure*}[!ht]
    \centering
    \includegraphics[width=1\textwidth]{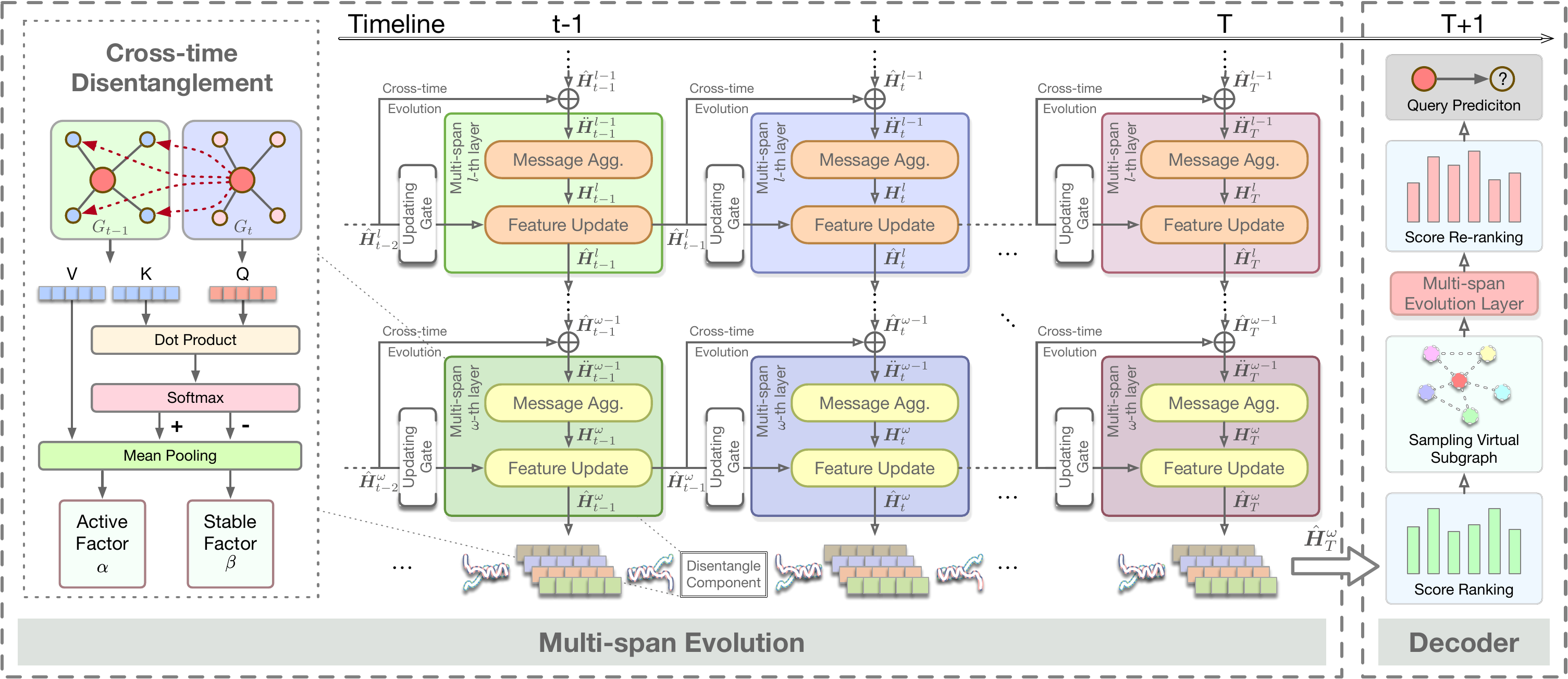}
    \caption{The Overall Architecture of DiMNet.}
    \label{fig:framework}
\end{figure*}

Evolving the learning of TKG subgraph sequences can better capture the dynamic process of semantic changes. 
Integrating the interaction of internal structural information among subgraphs can avoid the independence of subgraph modeling.
% allowing local structures modeling to more directly involve the structures of historical subgraphs.
As illustrated in Figure~\ref{fig:framework},
we equip a \textbf{Multi-span Evolution} strategy in DiMNet for TKGs to enable the local structure to perceive the historical neighbor information of nodes during the evolution of the subgraph sequence. Through the multi-span approach, historical neighbor features can assist in the fine-grained updating of nodes at the current timestamp.
We hope to involve historical neighbors with the same hop when the central node aggregates current different hop neighbors based on the graph neural network, as shown in Figure~\ref{fig:historical-neighbor}.
From a high-level perspective, DiMNet integrates the updating process of historical neighbors into the local subgraph structure modeling process.

To maximize the modeling of dynamic change patterns in historical subgraph sequences, we integrate a cross-time \textbf{Disentangle Component} into the gaps between subgraph modeling. 
It guides how evolved information influences the subsequent learning of historical subgraphs and how historical neighbors affect the subsequent central nodes. 
It adaptively disentangles the semantic changes of nodes in adjacent subgraphs, separating active and stable features of nodes, and uses them for dynamic control of historical information in the subsequent multi-span evolution process, achieving better temporal modeling of TKGs.

As historical subgraphs evolve node semantics over time, the evolved node representations implicitly contain temporal and semantic features. 
Existing methods, such as \cite{li2021temporal,zhang2023learning}, directly score the final representations to complete future facts. 
However, due to the unknown topology of subgraphs at future timestamps, there might be a distribution shift between the inferred future facts and the actual facts. 
We design a decoder specifically for multi-span evolution. It constructs an \textit{Inference Virtual Subgraph} based on the query, then revisits the multi-span evolved history to make a query-aware final score.

\subsection{Multi-span Evolution Backbone}\label{sec:multispan}

The essence of evolutionary learning on TKGs is to progressively learn the structural information of subgraphs at each timestamp and model the semantic changes of nodes across the subgraph sequence. 
Formally, given a historical subgraph sequence $\{G_{t-m+1},..., G_t\}$, the goal of the evolution backbone is to model the subgraphs along the historical timeline into a sequence of node embeddings $\{\hat{\bm{H}}_{t-m+1},..., \hat{\bm{H}}_t\}$.

In DiMNet, the evolutionary component adopts a GNN-based approach to model the structure of subgraphs at each timestamp. 
% Specifically, we employ $\omega$-layers message passing to capture the relationship, thereby obtaining the semantic information of the nodes. 
To smoothly incorporate historical neighbor information from different hops, we employ an $\omega$-layers multi-span iteration to integrate intermediate features from previous subgraphs. 
Each layer is designed in two parts: \textbf{message aggregation} and \textbf{feature update}.

\subsubsection{Message Aggregation Process}
% We start with the aggregation process. 
For a subgraph at timestamp $t$,  for an object node $o$ at the $l$-th layer ($l \in [1,\omega]$) is as follows:
\begin{equation}\label{eq:agg}
    \bm{h}_{o,t}^l = 
    % \sigma 
    % \Bigg(
    \mathtt{Agg}
    \bigg(
    \Big\{\underbrace{\bm{W}_{nbr}^l(\ddot{\bm{h}}_{s,t}^{l-1}\oplus \bm{r}^l)\big| (s,r,o)\in G_t}_{Neighbor Message}\Big\}
    \uplus
    \Big\{\underbrace{\bm{W}_{sf}^l\ddot{\bm{h}}_{o,t}^{l-1}}_{Self-loop}\Big\}
    \bigg).
    % \Bigg),
\end{equation}
The central node $o$ considers neighbor features and self-loop features during the aggregation.
The term $\bm{r}^l \in \mathbb{R}^{d}$ denotes the relation type representation between $s$ and $o$. It is important to note that, considering the different contributions of edges to semantic evolution at different layers, we process the relation embedding in a layer-specific manner using a linear function, i.e., $\bm{r}^l=\bm{W}_\mathtt{REL}^l \bm{r}+\bm{b}_\mathtt{REL}^l$, where $\bm{r}\in\bm{\mathcal{R}}$ is a learnable original relation representation\footnote{Here we draw on the method mentioned in \cite{lee2023ingram} to construct a relation affinity matrix to guide the initialization of relation embeddings, which will not be elaborated here.}.
$\bm{W}_{nbr}^l, \bm{W}_{sf}^l \in \mathbb{R}^{d \times d}$ are layer-specific transformation parameters for neighbor and self-loop features, respectively. $\oplus$ is defined as the element-wise summation operator. $\uplus$ denotes the operation of merging the two types of feature messages. $\mathtt{Agg}(\cdot)$ represents the message aggregation method, for which we use the widely adopted Principal Neighborhood Aggregation (PNA) proposed in \cite{corso2020principal}, which leverages multiple aggregators (namely mean, maximum, minimum, and standard deviation) to learn joint features.
% For $\ddot{\bm{h}}_{s,t}^{l-1}$ and $\ddot{\bm{h}}_{o,t}^{l-1}$, they contain the updated features of $s$ and $o$, respectively, at the current timestamp $t$ from the previous layer $l-1$ as well as the updated features from the same layer $l$ at the previous timestamp $t-1$, as follows:
For $\ddot{\bm{h}}_{s,t}^{l-1}, \ddot{\bm{h}}_{o,t}^{l-1} \in \mathbb{R}^{d}$, they are derived from the previously obtained final updated features, as follows:
\begin{equation}
    \ddot{\bm{H}}_t^{l-1} = \hat{\bm{H}}_t^{l-1} \oplus \bm{W}_{\mathtt{CE}}^{l} \hat{\bm{H}}_{t-1}^{l}.
\end{equation}
Here, we use $\ddot{\bm{H}}_*^* \in \mathbb{R}^{|\mathcal{V}| \times d}$ with uppercase to denote the representations of the entire set of nodes $\mathcal{V}$. For $s,o \in \mathcal{V}$, there is $\ddot{\bm{h}}_{s,t}^{l-1}, \ddot{\bm{h}}_{o,t}^{l-1} \in \ddot{\bm{H}}_t^{l-1}$. 
$\hat{\bm{H}}_*^{*}$ represents the final updated features, which will be introduced in Sec.~\ref{sec:update}. $\bm{W}_{\mathtt{CE}}^{l} \in \mathbb{R}^{d \times d}$ is a trainable transformation matrix used for the weighted summation of updated features from two different timestamps to achieve $\bm{\mathtt{C}}$ross-time $\bm{\mathtt{E}}$volution ($\mathtt{CE}$).
Through the above observations, we can conclude that the input of each layer is generated by: the output from the same layer at the previous timestamp, and the output from the previous layer at the current timestamp.

% $\ddot{\bm{h}}_{s,t}^{l-1}$ represents the updated features of the neighbor node $s$ at $(l-1)$-th layer, which also involves the evolutionary state from historical timestamps. This will be discussed in more detail later. 
% In the self-loop part, it accounts for the previous layer's state of the central node $o$ and the evolutionary information from historical timestamps. 

\subsubsection{Feature Update after Aggregation}\label{sec:update}

After completing the feature aggregation described in Eq.~\ref{eq:agg}, $\bm{h}_{o,t}^l$ is expected to have captured the subgraph and historical $l$-hop neighbor information for node $o$, and the same applies to other nodes $o' \in \mathcal{V}$. Similarly, we use $\bm{H}_{t}^l \in \mathbb{R}^{|\mathcal{V}| \times d}$ to represent the features obtained for the entire set of nodes at timestamp $t$ after $l$-th layer aggregation.
To better capture the temporal pattern of the subgraph sequence and non-trivial influence of historical neighbors on the semantic evolution process, we employ a multi-span gating mechanism to assist in updating the node semantics based on the aggregation process, as follows:

\begin{equation}\label{eq:update}
    \hat{\bm{H}}_t^l = \bm{U}_t^l \otimes \bm{H}_t^l + (1-\bm{U}_t^l)\otimes \hat{\bm{H}}_{t-1}^l.
\end{equation}
Eq.~\ref{eq:update} describes how to obtain the final updated representation $\hat{\bm{H}}_t^l \in \mathbb{R}^{|\mathcal{V}| \times d}$ of nodes using the aggregated features $\bm{H}_{t}^l$ after $l$-th layer aggregation. Here, $\otimes$ is defined as the element-wise multiplication operator. $\bm{U}_t^l \in \mathbb{R}^{|\mathcal{V}| \times d}$ represents the gating weights at timestamp $t$, which are obtained by 
\begin{equation}\label{eq:updategate}
    \bm{U}_t^l = \sigma \big( \bm{W}_\mathtt{UG}^l \bm{\mathcal{A}}_{t-1} + \bm{b}_\mathtt{UG}^l \big),
\end{equation}
where $\bm{W}_\mathtt{UG}^l \in \mathbb{R}^{d \times d}$ and $\bm{b}_\mathtt{UG}^l \in \mathbb{R}^{d}$ are the trainable weights and bias parameters in the $\bm{\mathtt{U}}$pdating $\bm{\mathtt{G}}$ate ($\mathtt{UG}$) computation process, respectively. As for $\bm{\mathcal{A}}_{t-1}$, it represents an active factor used to guide the influence of historical neighbors on feature update, that can be disentangled during the evolution process, which will be introduced in Sec.~\ref{sec:disentangle}.

We can see that $\bm{U}_t^l$, $\bm{W}_\mathtt{UG}^l$, and $\bm{b}_\mathtt{UG}^l$ are all layer-specific, ensuring that the influence of updated features from different historical layers on the current timestamp's update is unique. This multi-span strategy aligns with our goal by modeling the cross-time same-layer feature transmission, allowing the central node to perceive the semantic changes of historical neighbors.

\subsubsection{Initialization of the Starting Layer and Timestamp}

In the training phase, we initialize the trainable embeddings for the set of nodes and relations, denoted as $\bm{\mathcal{V}} \in \mathbb{R}^{|\mathcal{V}| \times d}$ and $\bm{\mathcal{R}} \in \mathbb{R}^{|\mathcal{R}| \times d}$, respectively. For $\forall t > 0$, which means for timestamps other than the starting timestamp of the subgraph sequence, the input for the $1$-st layer is derived as follows:
\begin{equation}
    \hat{\bm{h}}_{o,t}^0 = g\Big(\bm p_{o,t}\ \big|\big|\ \mathtt{MP}\{\bm{\mathcal{R}}_{\mathcal{K}(o)}\}\Big),
\end{equation}
where ${\mathcal{K}(o)}$ represents the edges that $o$ serve as objects, $\mathtt{MP}\{\cdot\}$ denotes the mean pooling operation performed on relation embeddings corresponding to ${\mathcal{K}(o)}$, $||$ represents the vector concatenation operation, and $g$ is a 2-layer fully connected network used to reduce the dimensionality to $\mathbb{R}^{d}$. And $\bm{p}_{o,t} \in \bm{P}_t$ is determined as follows:
\begin{equation}\label{eq:init>0}
    \bm{P}_t = \bm{I}_t \otimes \bm{\mathcal{V}} + (1-\bm{I}_t)\otimes \hat{\bm{H}}_{t-1}^\omega,
\end{equation}
\begin{equation}\label{eq:init>0gate}
    \bm{I}_t = \sigma \big( \bm{W}_\mathtt{IG} \bm{\mathcal{B}}_{t-1} + \bm{b}_\mathtt{IG} \big).
\end{equation}
Similar to Eq.~\ref{eq:update} \& Eq.~\ref{eq:updategate}, a gating mechanism parameterized by $\bm{\mathtt{I}}$nitialization $\bm{\mathtt{G}}$ate weight $\bm{W}_\mathtt{IG}$ is adopted here. $\hat{\bm{H}}_{t-1}^\omega$ represents the features obtained after the final $\omega$-th layer update at $t-1$, ensuring that while evolving across same-layer, the sequential temporal characteristics of the structure are also preserved.
And $\bm{\mathcal{B}}_{t-1}$, in contrast to $\bm{\mathcal{A}}_{t-1}$ in Eq.~\ref{eq:updategate}, represents a stable factor used to guide the initialization of the layer, which will also be introduced in Sec.~\ref{sec:disentangle}.
For $t=0$, $\bm{P}_t \leftarrow \bm{\mathcal{V}}$, and $\bm{\mathcal{A}}_{t-1}, \bm{\mathcal{B}}_{t-1} \leftarrow \vec{0}$.

% xERTE~\cite{han2020explainable} assumes that relations do not evolve and constructs an inference graph by sampling nodes from historical subgraphs, learning time-aware entity embeddings on it, yet neglecting complex relationship patterns.

\subsection{Cross-time Disentanglement}\label{sec:disentangle}

Since node semantic changes occur along with the changes in subgraph structures over time, here we design a cross-time disentangle component to learn how node semantics change after each subgraph structure,
% And it can be used to further guide the influence of historical evolutionary states on node semantic updates during the multi-span evolution process.(this can be removed)
utilizing the updated features at two adjacent timestamps, $\hat{\bm{H}}_t^\omega$ and $\hat{\bm{H}}_{t-1}^\omega$, where $\omega$ denotes the final layer of aggregation in the evolution. 
We aim to disentangle a pair of mutually exclusive factors, namely the \textbf{active factor} and the \textbf{stable factor}, to represent the activity and smoothness of the semantic changes of nodes between timestamp $t$ and $t-1$. 

Considering that the semantic changes of $\hat{\bm{h}}_{o,t}^\omega$ relative to $\hat{\bm{h}}_{o,t-1}^\omega$ are due to the differences in the neighbor at timestamp $t$ compared to $t-1$.
Therefore, we start from the updated feature of a central node $o$ at timestamp $t$ and the related $1$-hop neighbors $\mathcal{N}_{o,t-1}$ at timestamp $t-1$, since 1-hop neighbors have already collected sufficient neighbor features in $\hat{\bm{H}}_{t-1}^\omega$, where $\mathcal{N}_{o,t-1}=\{s|(s,r,o)\in G_{t-1}\}$.
Specifically, we design a mutually exclusive attention mechanism to disentangle the adjacent updated states. 
For $\forall s \in \mathcal{N}_{o,t-1}$ and $(s,r,o) \in G_{t-1}$, we calculate the attention vectors following\footnote{During the disentangling process, self-loop edges are added to $G_{t-1}$, and an additional representation for the self-loop edge type is learned, ensuring that isolated nodes at $t-1$ can also be properly disentangled.}:
\begin{align}
    \bm{Q}_o^t &= \bm{W}_\mathtt{Q}(\hat{\bm{h}}_{o,t}^\omega\ ||\ \bm r), \\
    \bm{K}_s^t &= \bm{W}_\mathtt{K}(\hat{\bm{h}}_{s,t-1}^\omega\ ||\ \bm r), \\
    \bm{V}_s^t &= \bm{W}_\mathtt{V}(\hat{\bm{h}}_{s,t-1}^\omega),
\end{align}
where $\hat{\bm{h}}_{s,t-1}^\omega$ represents the updated feature of node $o$'s neighbor $s$ at timestamp $t-1$. $\bm{W}_\mathtt{Q},\bm{W}_\mathtt{K} \in \mathbb{R}^{2d \times d}$, $\bm{W}_\mathtt{V} \in \mathbb{R}^{d \times d}$ denotes the weight matrices for query, key and value vector, respectively. Note that in the computation of $\bm{Q}_o^t$ and $\bm{K}_s^t$, we concatenate the relation embedding $\bm r$ to distinguish the impact of different edge types. For simplicity, we omit the bias terms here. Then, we can calculate the attention score between the neighbor $s \in \mathcal{N}_{o,t-1}$ and the central node $o$ through the dot product operation, as follows:
\begin{equation}
    e_{<s,o>} = \frac{\bm{Q}_o^t \cdot \bm{K}_s^t}{\sqrt{d}}.
\end{equation}
The positive and negative forms of $e_{<s,o>}$ are passed through the \textit{softmax} function to thus obtain two inversely proportional normalized scores, following:
\begin{align}
    \eta_{s,o}^t &= \frac{\exp(e_{<s,o>})}{\sum_{u \in \mathcal{N}_{o,t-1}} \exp(e_{<u,o>})}, \\
    \overline{\eta}_{s,o}^t &= \frac{\exp(-e_{<s,o>})}{\sum_{u \in \mathcal{N}_{o,t-1}} \exp(-e_{<u,o>})}.
\end{align}
Here, we denote $\eta$ as the active score and $\overline{\eta}$ as the stable score. Intuitively, a neighbor with higher $\eta$ will have lower $\overline{\eta}$. Finally, we perform mean pooling on all the neighbor features $\bm{V}_s^t$ based on the two scores to obtain a pair of mutually exclusive
factors, namely active factor $\bm{\alpha}$ and stable factor $\bm{\beta}$:
% \begin{align}
%    \bm{\alpha}_t &= \mathtt{MP}\{\eta_{u,o}^t \bm{V}_u^t\ |\ u\in \mathcal{N}_{o,t-1}\}, \\
%    \bm{\beta}_t &= \mathtt{MP}\{\overline{\eta}_{u,o}^t \bm{V}_u^t\ |\ u\in \mathcal{N}_{o,t-1}\}.
% \end{align}
\begin{align}
    \label{eq:activefactor}
    \forall u\in \mathcal{N}_{o,t-1}, \ \
    \bm{\alpha}_t^o &= \mathtt{GRU}(\bm{\alpha}_{t-1}^o, \mathtt{MP}\{\eta_{u,o}^t \bm{V}_u^t\}),\\
    \forall u\in \mathcal{N}_{o,t-1}, \ \
    \bm{\beta}_t^o &= \mathtt{MP}\{\overline{\eta}_{u,o}^t \bm{V}_u^t\}.
\end{align}
Note that $\bm{\alpha}_t^o \in \bm{\mathcal{A}}_{t}$ and $\bm{\beta}_t^o \in \bm{\mathcal{B}}_{t}$.
% As mentioned in Sec.~\ref{sec:multispan}, they are used to guide\footnote{In practice, we also add trainable weights to active factor $\bm{\mathcal{A}}_{t}$ to enhance the learning of the activity in node semantic changes over time.} the dynamic gating parameters $\bm{U}_*$ and $\bm{I}_*$ (corresponding to Eq.~\ref{eq:updategate} and Eq.~\ref{eq:init>0gate}, respectively).
As mentioned in Sec.~\ref{sec:multispan}, they are used to guide the dynamic gating parameters $\bm{U}_*$ and $\bm{I}_*$ (corresponding to Eq.~\ref{eq:updategate} and Eq.~\ref{eq:init>0gate}, respectively).
Here we see that the calculation of $\bm{\alpha}_t^o$ incorporates a $\mathtt{GRU}$ iterative process. This endows the active factor with temporal characteristics, allowing it to learn the temporal patterns of activity along the subgraph sequence and thus providing better temporal features to guide semantic evolution.

\subsection{Decoder and Inference}
After the subgraph sequence of a TKG undergoes multi-span evolution, the complex evolutionary features of node semantics are captured. 
% By employing a multi-span strategy, nodes can perceive historical neighbor information in a more fine-grained manner during semantic change modeling. 
Based on this, we design an inference scheme specifically for the multi-span evolution encoding.

\subsubsection{Score Function}
Given a historical subgraph sequence ending with $G_T$, where $T$ is the final timestamp in the given historical sequence, the final evolution result $\hat{\bm{H}}_T^\omega$ retains all the semantic and temporal information from the subgraph sequence. 
Therefore, for the query $(s,r,?)$ at timestamp $T+1$, we consider using $\hat{\bm{H}}_T^\omega$ to decode and calculate the probability of interaction between subject node $s$ and object candidates $\forall o \in \mathcal{V}$ under the relation $r$.

Specifically, we use the widely adopted scoring function ConvTransE~\cite{shang2019end} to score the missing object node following Eq.~\ref{eq:score}, where $\hat{\bm{h}}_{s,T}^\omega, \hat{\bm{h}}_{o,T}^\omega \in \hat{\bm{H}}_T^\omega$, $\sigma$ is sigmoid function.
\begin{equation}\label{eq:score}
    \phi_{T+1}(o|s,r) = \sigma\Big(\hat{\bm{h}}_{o,T}^\omega \mathrm{ConvTransE}(\hat{\bm{h}}_{s,T}^\omega,\bm{r})\Big)
\end{equation}

\subsubsection{Inference by Sampling Virtual Subgraph} 
We note that the topologies between subgraphs are mutually independent. This means that when predicting the subgraph at a future timestamp, solely relying on the updated features $\hat{\bm{H}}_T^\omega$ obtained at timestamp $T$ is insufficient, as DiMNet does not incorporate query-related features during the encoding phase. Therefore, inspired by the query-aware method~\cite{dong2023adaptive, dong2024temporal}, we sample a few edges using the scoring results derived by Eq.~\ref{eq:score} to form a virtual graph, which is used for the final scoring inference, as follows:
\begin{equation}
    G_{\mathtt{INF}}^{T+1}=\bigg\{(s,r,o_i)\Big|o_i \in \text{Top-}k\big[\Phi_{T+1}(s,r)\big]\bigg\},
\end{equation}
where $\Phi_{T+1}(s,r)$ represents the scores of all candidate tail entities corresponding to the query $(s,r,?)$, i.e., $\phi_{T+1}(o_i|s,r) \in \Phi_{T+1}(s,r)$. $s$ and $r$ are the subject and relation of the query at timestamp $T+1$. $\text{Top-}k[\cdot]$ denotes sampling top $k$ candidates with the highest scores.

Based on the virtual graph $G_{\mathtt{INF}}^{T+1}$, we perform multi-span evolution (Eq.~\ref{eq:agg}-\ref{eq:init>0gate}) once again to obtain $\hat{\bm{H}}_{T+1}^\omega$, which is used to fed into the scoring function again to obtain the final scoring results:
\begin{equation}\label{eq:score2}
    \phi_{T+1}^\prime(o|s,r) = \sigma\Big(\hat{\bm{h}}_{o,T+1}^\omega \mathrm{ConvTransE}(\hat{\bm{h}}_{s,T+1}^\omega,\bm{r})\Big).
\end{equation}

% \noindent Note that the used decoder is shared with the mentioned above, making a more reasonable decoded score.

\subsection{Learning and Optimization}
Predicting missing facts can be viewed as a multi-label learning task. Therefore, based on the final scoring results, the prediction loss is formalized as follows:
\begin{equation}
    \mathcal{L}_{pred} = 
    \sum_{t=1}^{|\mathcal{T}|}
    \sum_{(s,r,o)\in G_t} 
    \sum_{u\in \mathcal{V}} 
    y_t^{u} \log\phi_t^\prime(u|s,r),
\end{equation}
where $|\mathcal{T}|$ represents the number of timestamps in the training set. And $y_t^{u}$ is the ground truth label, where the value is $1$ indicates that the fact occurs and $0$ otherwise.

Additionally, during the evolution process, we disentangle the updated features of adjacent timestamps to obtain the active factor $\bm{\alpha}$ and the stable factor $\bm{\beta}$. To ensure that the stable factor captures smooth features during training, we introduce an additional distance constraint loss:
\begin{equation}
    \mathcal{L}_{dis} = 
    \sum_{t=1}^{m}
    \sum_{u\in \mathcal{V}} 
    \Big(1 - \text{CosineSim}[\bm{\beta}_{t-1}^u, \bm{\beta}_{t}^u]\Big).
\end{equation}
Note that, for simplicity, we omit the outer summation symbol over the absolute timestamps in the training set, and use $t$ to denote the relative timestamps in the input historical sequence, with $m$ representing the length of the historical sequence. $\text{CosineSim}[\cdot]$ denotes the computation of the cosine similarity between vectors, used to ensure that the stable factors obtained at adjacent timestamps are similar.
Based on the above discussion, the final loss can be expressed as:
\begin{equation}
    \mathcal{L} = \mathcal{L}_{pred} + \mathcal{L}_{dis}.
\end{equation}

%% file: 5.exp.tex
% \clearpage

\input{table/overall}

\section{Experiments}

\subsection{Experimental Setup}
\subsubsection{Datasets}

We adopt four widely used datasets: ICEWS14~\cite{han2020explainable}, ICEWS05-15~\cite{li2021temporal}, ICEWS18~\cite{jin2020Renet}, and GDELT~\cite{2013GDELT}. 
Specifically, ICEWS14, ICEWS05-15, and ICEWS18 are subsets generated from the Integrated Crisis Early Warning System~\cite{DVN/28075_2015}, which contains political events with specific timestamps. 
GDELT is derived from the Global Database of Events, Language, and Tone.
We follow the data processing strategies mentioned in~\cite{jin2020Renet, dong2024temporal}, which split the dataset into train/valid/test by timestamps that (timestamps of the train) $<$ (timestamps of the valid) $<$ (timestamps of the test). Detailed dataset statistics are provided in Table~\ref{tab:dataset}.

\input{table/dataset}

\subsubsection{Evaluation Metrics}
To evaluate the performance of DiMNet on TKG reasoning, we also choose the widely used task of link prediction on future timestamps corresponding to the query. 
We use Mean Reciprocal Rank (MRR) and Hits@\{1, 3, 10\} as performance evaluation metrics, which are widely reported by previous methods. 
MRR calculates the average of the reciprocal ranks of the ground truth for all queries. 
Hits@k measures the proportion of times that the true candidates appear in the top-k of the ranked candidates. 
Some existing methods use filtered metrics that exclude all valid quadruples in the training, validation, or test sets. 
However, this setting has been shown to be unsuitable for TKG reasoning tasks~\cite{dong2024temporal,han2021learning,dong2023adaptive}. 
Therefore, we report the more reasonable time-aware filtered metrics, which only filter out valid facts at the query timestamp from the rank list.

\subsubsection{Baseline Methods}
To fully demonstrate the effectiveness of DiMNet, we conduct a comprehensive comparison of our proposed model with up-to-date baselines that include traditional KG reasoning methods, as well as interpolated and extrapolated TKG reasoning methods:
(1) Traditional KG reasoning model does not consider temporal features, including DistMult\cite{yang2014embedding}, ComplEx~\cite{2016Complex}, ConvE~\cite{2017Convolutional}, and RotatE~\cite{2019RotatE}. 
(2) Interpolated TKG Reasoning methods include TTransE~\cite{2018Deriving}, TA-DistMult~\cite{A2018Learning}, DE-SimplE~\cite{2020Diachronic}, and TNTComplEx~\cite{2020Tensor}. 
(3) Extrapolated TKG Reasoning methods include TANGO-Tucker~\cite{han2021learning}, TANGO-DistMult~\cite{han2021learning}, CyGNet~\cite{zhu2021learning}, RE-NET~\cite{jin2020Renet}, RE-GCN~\cite{li2021temporal}, TITer~\cite{sun2021timetraveler}, xERTE~\cite{han2020explainable}, CEN~\cite{li-etal-2022-complex}, GHT~\cite{sun2022graph}, DaeMon~\cite{dong2023adaptive}, and TiPNN~\cite{dong2024temporal}.

\subsection{Implementation Details}
For all datasets, the embedding dimension $d$ is set to 128, and we use Adam~\cite{kingma2014adam} for parameter learning with a learning rate of ${1e-3}$. 
For the multi-span evolution, we perform a grid search on the history length $m$ and present overview results with the lengths 10, 2, 10, 5, corresponding to the datasets ICEWS14, ICEWS05-15, ICEWS18, and GDELT in Table \ref{tab:overall_results}, of which parameter sensitivity analysis is shown in Figure \ref{fig:history_length}. 
The number of evolution layers $\omega$ in each timestamp is set to 3 for the ICEWS14, ICEWS18, GDELT datasets, and 1 for the ICEWS05-15 dataset. Layer normalization and shortcuts are applied for the message aggregation process of each layer. The adopted activation function is $RReLU$.
For the cross-time disentanglement, we incorporate a multi-head design into the attention mechanism to learn more complex features. The number of heads is set to 4 for ICEWS14 and ICEWS18, and to 1 for ICEWS05-15 and GDELT. 
The sampling number $k$ is set to 50 for all datasets, during the construction of the virtual subgraph for scoring.
The maximum epoch in the training stage has been set to 60.
All experiments were conducted with TESLA A100 GPUs.

\subsection{Experimental Results}
To demonstrate the effectiveness of DiMNet, we conducted experiments on four widely used benchmark datasets to show the overall performance of DiMNet in TKG reasoning tasks, as shown in Table~\ref{tab:overall_results}. Values highlighted in \textbf{bold} represent the best results on each dataset. From the overall results, we find that DiMNet outperforms all baseline methods and achieves state-of-the-art performance on the four TKG datasets. Specifically, it surpasses the existing best methods in terms of MRR by 8.4\%, 22.7\%, 6.1\%, and 3.6\% on ICEWS14, ICEWS05-15, ICEWS18, and GDELT datasets, respectively.

Compared to traditional KG reasoning methods (listed in the first block of Table~\ref{tab:overall_results}) and interpolated TKG reasoning methods (second block), DiMNet considers the temporal characteristics of facts and employs an evolutionary modeling strategy to maximize the capture of historical semantic evolution features. This gives it a natural advantage in TKG reasoning, showcasing excellent performance in completing future missing facts.
In comparison with existing extrapolated TKG reasoning methods (third block), DiMNet still demonstrates outstanding performance. Unlike previous methods, in modeling the semantic evolution of historical subgraph sequences, DiMNet incorporates a multi-span strategy that allows the model to perceive intermediate features of historical neighbor semantic updates while capturing local structural semantics, facilitating the learning of multi-span semantic evolution, which is overlooked by existing evolution-based methods~\cite{li2021temporal, liu2023retia, dong2023adaptive}. The disentangle component in DiMNet adaptively separates the active and stable features of nodes during the semantic change process, guiding the influence of historical neighbor features on subsequent semantic updates in the multi-span evolution process, thereby maximizing the capture of semantic change patterns. Furthermore, the specifically designed decoder synergizes with historical evolutionary modeling to enhance the performance in completing future missing facts.

\input{table/ablation}

\subsection{Ablation Study}

To verify the impact of each component, we conducted an ablation study on all datasets. The experimental results of DiMNet variants are shown in Table~\ref{tab:ablation}.

\textit{Multi-span evolution strategy} (denoted as w/o Multi-span): Note that we don't remove the entire evolution strategy but rather disable the perception of historical neighbor features during the local structural semantics capturing. This way, the basic temporal features can still be retained, thereby maximizing the impact of the multi-span strategy. From the results, it is evident that without the perception of historical neighbors, the performance of DiMNet significantly drops as it can no longer capture the fine-grained multi-span updates and evolution.
\textit{Disentangle component} (denoted as w/o Disentangle): We replaced the active factor $\bm{\mathcal{A}}$ and stable factor $\bm{\mathcal{B}}$ calculated by the disentangle component with zero vectors, removed the $\mathtt{GRU}$ iterative process (in Eq.~\ref{eq:activefactor}), and deleted the $\mathcal{L}_{dis}$ loss term. The results show that without the dynamic disentangled features, DiMNet also suffers significant performance loss because the disentangle component adaptively separates complex changing features while providing updated guidance for multi-span evolution.
\textit{Combined impact of multi-span and disentangle components} (denoted as w/o Multi-span \& Disentangle): When both the multi-span mechanism and the disentangle component are removed, DiMNet almost degrades to the stacking of $m*\omega$ GNN layers, of which the results are still not satisfactory.
\textit{Sampling virtual subgraph} (denoted as w/o $G_{\mathtt{INF}}$): We also attempted to remove the inference strategy using the virtual subgraph for re-scoring, and the results were similarly poor. This aligns with our intuition and further proves the effectiveness of DiMNet’s design.

\subsection{Parameter Study}

\begin{figure}[!t]
\centering
\subfigure[ICEWS14]{
\includegraphics[width=0.49\linewidth]{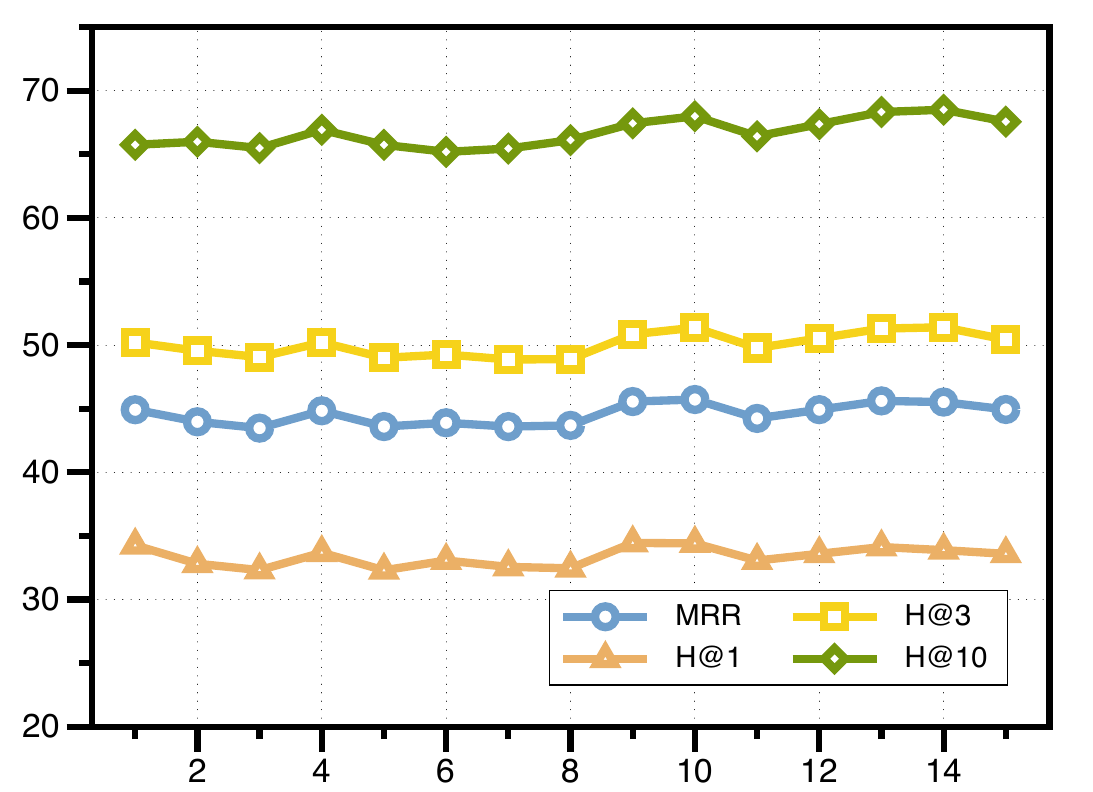}
}
\hspace{-4mm}
\subfigure[ICEWS18]{
\includegraphics[width=0.49\linewidth]{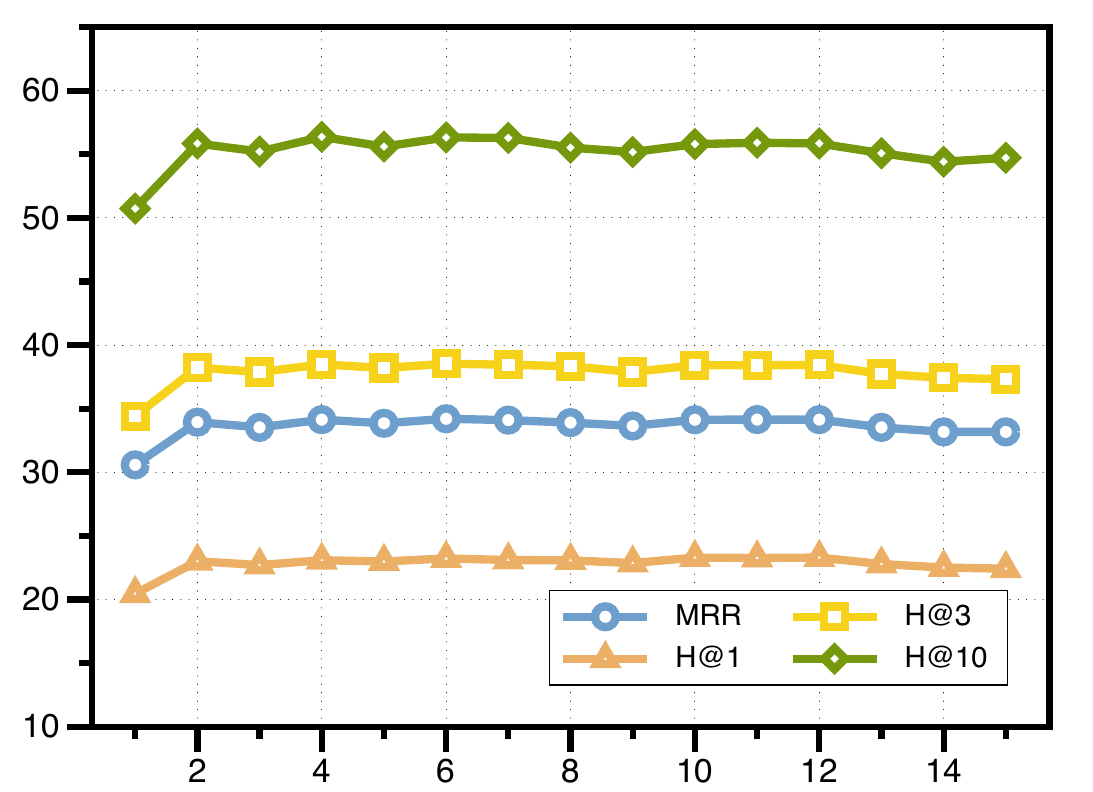}
}
\vspace{-4mm}
\caption{Performance on Different Sequence Length $m$.}
\label{fig:history_length}
\vspace{-4mm}
\end{figure}

\begin{figure}[!t]
\centering
\subfigure[ICEWS14]{
\includegraphics[width=0.49\linewidth]{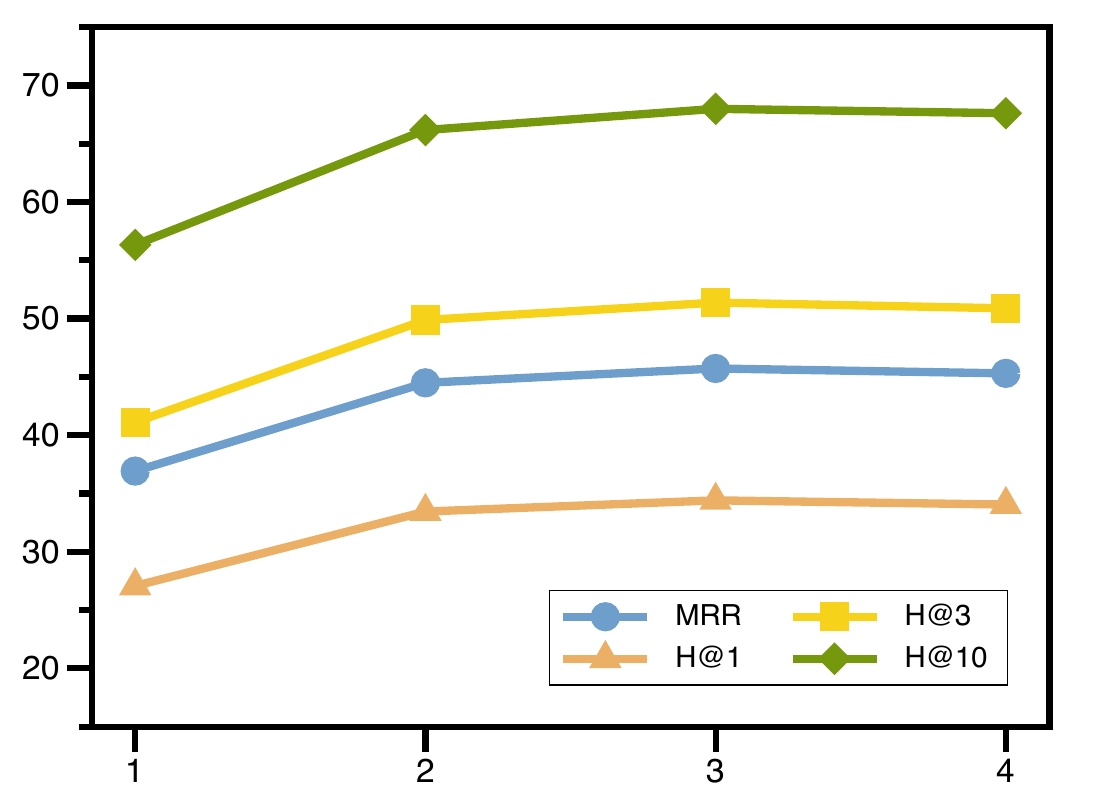}
}
\hspace{-4mm}
\subfigure[ICEWS18]{
\includegraphics[width=0.49\linewidth]{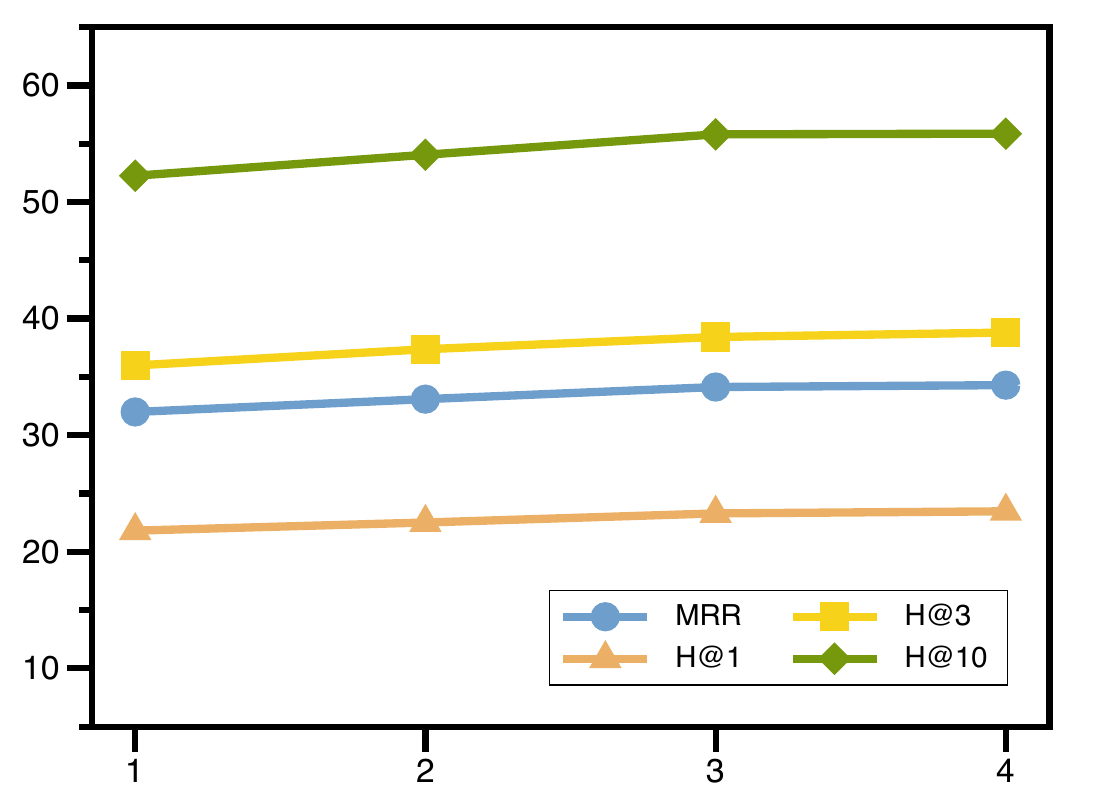}
}
\vspace{-4mm}
\caption{Performance on Different GNN Layer Number $\omega$.}
\label{fig:lynum}
\vspace{-4mm}
\end{figure}

To provide more insights on DiMNet's model parameters, we test the performance of different historical sequence lengths $m$, GNN layer numbers $\omega$, and virtual subgraph sampling numbers $k$ on ICEWS14 and ICEWS18 datasets.

\textit{Analysis of Sequence Length $m$}: We conduct experiments with different values of $m$, and the experimental results are shown in Figure~\ref{fig:history_length}. As seen in the figure, the historical sequence length $m$ in DiMNet does not significantly affect performance on both datasets. Especially for ICEWS18, the performance change tends to stabilize, demonstrating the robustness of DiMNet. In practice, to balance performance and complexity, we set the optimal number of $m$ to 10 for both ICEWS14 and ICEWS18.

\textit{Analysis of Layer Number $\omega$}: The number of layers directly impacts DiMNet's ability to model local neighborhood structures and perceive historical neighbor features during multi-span evolution. We experimented with different layer numbers to analyze the performance changes in DiMNet, as shown in Figure~\ref{fig:lynum}. The results indicate that ICEWS14 and ICEWS18 perform best with 3 layers. The number of layers has a relatively sensitive impact on performance for the ICEWS14 dataset, but the model performance tends to stabilize as the number of layers increases in both of the datasets.

\begin{figure}[!ht]
\centering
\subfigure[ICEWS14]{
\includegraphics[width=0.49\linewidth]{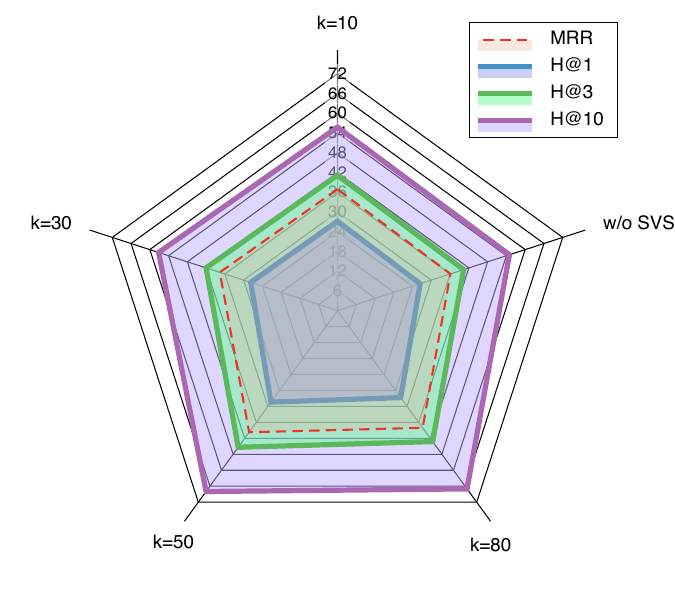}
}
\hspace{-4mm}
\subfigure[ICEWS18]{
\includegraphics[width=0.49\linewidth]{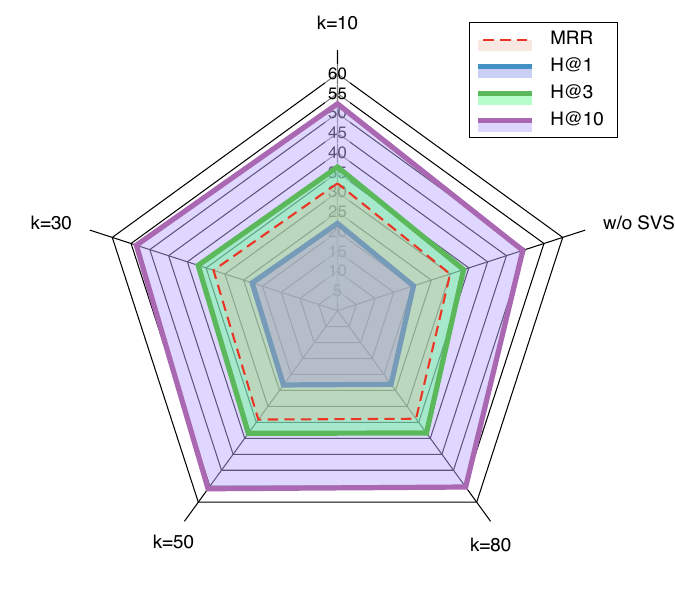}
}
\vspace{-1mm}
\caption{Performance on Different Virtual Subgraph Sampling Number $k$.}
\label{fig:topk}
\vspace{-1mm}
\end{figure}

\textit{Analysis of Sampling Number $k$}: Sampling the Virtual Subgraph helps mitigate the uncertainty of future topology during the inference stage. We experimented with different edge sampling numbers $k$, and the results are shown in Figure~\ref{fig:topk}, where w/o SVS indicates the removal of Sampling Virtual Subgraph. We can see that it significantly improves DiMNet's reasoning performance. However, for ICEWS14, when $k$ increases from 50 to 80, the model performance shows a declining trend. This is because the sampled subgraph is based on scoring, and a larger $k$ value introduces more erroneous noise, leading to poorer results.

%% file: table/overall.tex
\begin{table*}[!ht]
    \centering
    \small
    % \scriptsize
    % \footnotesize
    \setlength\tabcolsep{4pt}
    \caption{Overall Performance Comparison of Different Methods. Evaluation metrics are time-aware filtered MRR and Hits@\{1,3,10\}. All results are multiplied by 100. The best results are highlighted in \textbf{bold}.}
    \begin{tabular}{ccccccccccccccccc}

    \toprule
    \multirowcell{2.7}[0pt][c]{Model} &  \multicolumn{4}{c}{ICEWS14} & \multicolumn{4}{c}{ICEWS05-15} & \multicolumn{4}{c}{ICEWS18} & \multicolumn{4}{c}{GDELT} \\ 
    \cmidrule(lr){2-5}\cmidrule(lr){6-9}\cmidrule(lr){10-13}\cmidrule(lr){14-17}
    & MRR & H@1 & H@3 & H@10 & MRR & H@1 & H@3 & H@10 & MRR & H@1 & H@3 & H@10 & MRR & H@1 & H@3 & H@10 \\ 
    \midrule
    DistMult       & 16.16 & 11.42 & 17.94 & 25.30 & 17.95 & 13.12 & 20.71 & 29.32 & 11.51 & 7.03  & 12.87 & 20.86 & 8.68  & 5.58  & 9.96  & 17.13 \\
    ComplEx        & 21.28 & 14.49 & 23.11 & 35.20 & 32.63 & 24.01 & 37.50 & 52.81 & 22.94 & 15.19 & 27.05 & 42.11 & 16.96 & 11.25 & 19.52 & 32.35 \\
    ConvE          & 34.50 & 24.83 & 38.56 & 53.88 & 33.81 & 24.78 & 39.00 & 54.95 & 24.51 & 16.23 & 29.25 & 44.51 & 16.55 & 11.02 & 18.88 & 31.60 \\
    RotatE         & 20.88 & 10.26 & 23.90 & 44.03 & 24.71 & 13.22 & 29.04 & 48.16 & 12.78 & 4.01  & 14.89 & 31.91 & 13.45 & 6.95  & 14.09 & 25.99 \\
    \midrule
    TTransE        & 13.43 & 3.11  & 17.32 & 34.55 & 15.57 & 4.80  & 19.24 & 38.29 & 8.31  & 1.92  & 8.56  & 21.89 & 5.50  & 0.47  & 4.94  & 15.25 \\
    TA-DistMult    & 26.47 & 17.09 & 30.22 & 45.41 & 24.31 & 14.58 & 27.92 & 44.21 & 16.75 & 8.61  & 18.41 & 33.59 & 12.00 & 5.76  & 12.94 & 23.54 \\
    DE-SimplE      & 32.67 & 24.43 & 35.69 & 49.11 & 35.02 & 25.91 & 38.99 & 52.75 & 19.30 & 11.53 & 21.86 & 34.80 & 19.70 & 12.22 & 21.39 & 33.70 \\
    TNTComplEx     & 32.12 & 23.35 & 36.03 & 49.13 & 27.54 & 9.52  & 30.80 & 42.86 & 21.23 & 13.28 & 24.02 & 36.91 & 19.53 & 12.41 & 20.75 & 33.42 \\
    \midrule
    TANGO-Tucker   & 26.25 & 17.30 & 29.07 & 44.18 & 42.86 & 32.72 & 48.14 & 62.34 & 28.68 & 19.35 & 32.17 & 47.04 & 19.42 & 12.34 & 20.70 & 33.16 \\
    TANGO-DistMult & 24.70 & 16.36 & 27.26 & 41.35 & 40.71 & 31.23 & 45.33 & 58.95 & 26.65 & 17.92 & 30.08 & 44.09 & 19.20 & 12.17 & 20.40 & 32.78 \\
    CyGNet         & 32.73 & 23.69 & 36.31 & 50.67 & 36.81 & 26.61 & 41.63 & 56.22 & 24.93 & 15.90 & 28.28 & 42.61 & 18.48 & 11.52 & 19.57 & 31.98 \\
    RE-NET         & 38.28 & 28.68 & 41.43 & 54.52 & 43.32 & 33.43 & 47.77 & 63.06 & 28.81 & 19.05 & 32.44 & 47.51 & 19.62 & 12.42 & 21.00 & 34.01 \\
    RE-GCN         & 41.78 & 31.58 & 46.65 & 61.51 & 48.03 & 37.33 & 53.85 & 68.27 & 30.58 & 21.01 & 34.34 & 48.75 & 19.64 & 12.42 & 20.90 & 33.69 \\
    TITer          & 41.73 & 32.74 & 46.46 & 58.44 & 47.69 & 37.95 & 52.92 & 65.81 & 29.98 & 22.05 & 33.46 & 44.83 & 15.46 & 10.98 & 15.61 & 24.31 \\
    xERTE          & 40.79 & 32.70 & 45.67 & 57.30 & 46.62 & 37.84 & 52.31 & 63.92 & 29.31 & 21.03 & 33.40 & 45.60 & 18.09 & 12.30 & 20.06 & 30.34 \\
    CEN            & 42.17 & 32.10 & 47.59 & 61.43 & 46.84 & 36.38 & 52.45 & 67.01 & 30.84 & 21.23 & 34.58 & 49.67 & 20.18 & 12.84 & 21.51 & 34.10 \\
    GHT            & 37.40 & 27.77 & 41.66 & 56.19 & 40.31 & 29.99 & 45.04 & 60.51 & 29.16 & 18.99 & 33.16 & 48.37 & 20.13 & 12.87 & 21.30 & 34.19 \\
    DaeMon         & 40.68 & 31.53 & 45.58 & 56.73 & 44.50 & 35.55 & 49.64 & 60.75 & 31.85 & 22.67 & 35.92 & 49.80 & 20.73 & 13.65 & 22.53 & 34.23 \\
    TiPNN          & 41.79 & 32.76 & 46.92 & 58.80 & 45.35 & 36.27 & 50.58 & 62.16 & 32.17 & 22.74 & 36.24 & 50.72 & 21.17 & 14.03 & 22.98 & 34.76 \\
    
    \midrule
    \textbf{DiMNet} & \textbf{45.72} & \textbf{34.41} & \textbf{51.37} & \textbf{67.99} & \textbf{58.93} & \textbf{48.55} & \textbf{65.16} & \textbf{78.33} & \textbf{34.13} & \textbf{23.29} & \textbf{38.42} & \textbf{55.80} & \textbf{21.93} & \textbf{14.03} & \textbf{23.57} & \textbf{37.49} \\

    \bottomrule
    \end{tabular}
    
    \label{tab:overall_results}
    \vspace{-1mm}

\end{table*}

%% file: table/dataset.tex
\begin{table}[!ht]
    \centering
    % \small
    % \scriptsize
    % \footnotesize
    \setlength\tabcolsep{3.5pt}  
    \caption{Statistics of Datasets ($\mathcal{E}_{train}$, $\mathcal{E}_{valid}$, $\mathcal{E}_{test}$ denote the numbers of facts in training, validation, and test sets.).}
    \begin{tabular}{ccccccc}
    \toprule
    % \hline
       Datasets  & $|\mathcal{V}|$ & $|\mathcal{R}|$ & $\mathcal{E}_{train}$ & $\mathcal{E}_{valid}$ & $\mathcal{E}_{test}$ & $|\mathcal{T}|$\\
    \midrule
    ICEWS14 & 7,128 & 230 & 63,685 & 13,823 & 13,222 & 365 \\
    ICEWS05-15 & 10,488 & 251 & 368,868 & 46,302 & 46,159 & 4,017 \\
    ICEWS18 & 23,033 & 256 & 373,018 & 45,995 & 49,545 & 304 \\
    GDELT & 7,691 & 240 & 1,734,399 & 238,765 & 305,241 & 2,976 \\
    \bottomrule
    \end{tabular}
    
    \label{tab:dataset}
    \vspace{-4mm}
\end{table}

%% file: table/ablation.tex
\begin{table*}[!ht]
\centering
\small
% \scriptsize
% \setlength\tabcolsep{4.2pt}
% \footnotesize
\setlength\tabcolsep{3pt}
\caption{Experimental Results of Ablation Study.}
\begin{tabular}{lcccccccccccccccc}
    \toprule
    % \multirow{2.5}{*}{Model} &  \multicolumn{4}{c|}{ICEWS18} & \multicolumn{4}{c}{GDELT} \\ 
    \multirowcell{2.7}[0pt][c]{Model} &  \multicolumn{4}{c}{ICEWS14} & \multicolumn{4}{c}{ICEWS05-15} & \multicolumn{4}{c}{ICEWS18} & \multicolumn{4}{c}{GDELT} \\ 
    \cmidrule(lr){2-5}\cmidrule(lr){6-9}\cmidrule(lr){10-13}\cmidrule(lr){14-17}
    & MRR & H@1 & H@3 & H@10 & MRR & H@1 & H@3 & H@10 & MRR & H@1 & H@3 & H@10 & MRR & H@1 & H@3 & H@10 \\ 
    \midrule
    \textbf{DiMNet} & \textbf{45.72} & \textbf{34.41} & \textbf{51.37} & \textbf{67.99} & \textbf{58.93} & \textbf{48.55} & \textbf{65.16} & \textbf{78.33} & \textbf{34.13} & \textbf{23.29} & \textbf{38.42} & \textbf{55.80} & \textbf{21.93} & \textbf{14.03} & \textbf{23.57} & \textbf{37.49} \\
    w/o Multi-span & 40.75 & 30.16 & 45.64 & 61.62 & 51.17 & 40.80 & 56.97 & 70.88 & 30.74 & 20.67 & 34.57 & 50.94 & 20.99 & 13.06 & 22.55 & 37.03 \\
    w/o Disentangle & 34.34 & 24.90 & 38.16 & 52.99 & 53.22 & 43.14 & 59.05 & 72.01 & 33.60 & 22.88 & 37.97 & 54.91 & 20.51 & 12.64 & 22.56 & 36.23 \\
    w/o Multi-span \& Disentangle & 36.09 & 25.35 & 40.72 & 57.45 & 50.36 & 39.99 & 55.96 & 70.25 & 30.88 & 20.61 & 34.74 & 51.57 & 20.71 & 12.69 & 22.20 & 36.55 \\
    w/o $G_{\mathtt{INF}}$ & 36.10 & 26.42 & 40.33 & 54.96 & 45.45 & 35.15 & 50.93 & 65.22 & 30.02 & 20.25 & 33.61 & 49.37 & 19.81 & 12.54 & 21.10 & 33.92 \\

     \bottomrule
\end{tabular}

\label{tab:ablation}
\end{table*}

%% file: 2.rel.tex
\section{Related Work}

We briefly review the classic traditional KG reasoning methods and then introduce some recent related TKG reasoning methods.

Traditional KG reasoning methods aim to model static knowledge~\cite{ning2022graph,ning2025rethinking} and have seen significant advancements~\cite{2019RotatE, li-etal-2022-transher,xiao2022should,qiao2023semi}. There are some translation-based methods, such as TransE~\cite{bordes2013translating}, TransH~\cite{wang2014knowledge}, and TransR~\cite{lin2015learning}, view relations as translations of subject entities to object entities in the vector space. 
For the semantic matching-based methods, such as RESCAL~\cite{nickel2011three}, which measures the semantic matching between entities and relations through a tensor-based relational learning approach.
% and DistMult~\cite{yang2014embedding}, which simplifies RESCAL for efficiency by restricting relation matrices to diagonal matrices. 
Neural network-based methods, such as R-GCN~\cite{schlichtkrull2017modeling} and CompGCN~\cite{vashishth2019composition}, learn representations of entities and relations from the perspective of graph structural features.

To represent facts with temporal feature, recent literature has increasingly focused on learning TKGs. From the perspective of reasoning tasks, they are usually divided into interpolation and extrapolation reasoning. For interpolation, TTransE~\cite{2018Deriving} extends TransE~\cite{bordes2013translating} by binding timestamps to relations as translation features. HyTE~\cite{dasgupta2018hyte} links timestamps to their respective hyperplanes. TNTComplEx~\cite{2020Tensor}, building on ComplEx~\cite{trouillon2016complex}, treats TKGs as a 4th-order tensor and learns representations via canonical decomposition.
However, these methods are not particularly effective for predicting future facts~\cite{2020Diachronic, han2020dyernie, 2018Deriving, sadeghian2016temporal}, leading to the development of extrapolation methods. CyGNet~\cite{zhu2021learning} proposes a copy-generation mechanism to collect repeated events for query head entities and relations. RE-NET~\cite{jin2020Renet} utilizes sequence modeling and GCN to capture temporal and structural dependencies in TKG sequences. RE-GCN~\cite{li2021temporal} considers both subgraph structures and static properties, modeling them in an evolving manner. TANGO~\cite{han2021learning} employs neural ordinary differential equations for continuous-time reasoning. xERTE~\cite{han2020explainable} constructs an enclosing subgraph around the query through iterative sampling and attention propagation for reasoning. DaeMon~\cite{dong2023adaptive} proposes capturing query-aware temporal path features in sequential subgraphs to complete future missing facts. TiPNN~\cite{dong2024temporal} builds a history temporal graph from subgraph sequences and uses query-aware methods to learn relevant reasoning paths for queries.

As discussed earlier, there is still significant room for improvement in TKG reasoning methods. The proposed methods provide valuable solutions for learning semantic evolution patterns.

%% file: 6.con.tex
% \vspace{-2mm}
\section{Conclusion}
In this paper, we proposed a novel method DiMNet for TKG reasoning, which models the evolution of historical subgraph sequences from the perspective of semantic change. Based on message passing, we introduced a multi-span evolution strategy that enables interaction between the internal structures of subgraphs while modeling semantic evolution across historical subgraphs. To maximize the modeling of semantic change patterns, we designed a disentangle component to dynamically and adaptively separate the active and stable factors of node changes, guiding the update of node features in subsequent historical subgraphs. For multi-span evolution encoding over TKG, we designed an inference strategy based on sampling virtual subgraphs to help mitigate the uncertainty of future topology during the reasoning phase. Extensive experiments on four benchmark datasets demonstrate that DiMNet performs significant improvements, and achieves new state-of-the-art results.